\documentclass[11pt]{article}
\usepackage{eamt23}

\usepackage{times}
\usepackage{url}
\usepackage{latexsym}
\usepackage[small,bf]{caption} 
\usepackage{todonotes}

\usepackage{booktabs,siunitx,rotating}
\usepackage{times}
\usepackage{latexsym}
\usepackage{subcaption}

\usepackage[T1]{fontenc}

\usepackage[utf8]{inputenc}

\usepackage{microtype}

\usepackage{booktabs}
\usepackage{multicol,multirow}

\title{Investigating Lexical Sharing in Multilingual Machine Translation \\for Indian Languages}

\author{Sonal Sannigrahi \\
    Saarland University, Saarland Informatics Campus\\ Saarbrücken, Germany \\
  \texttt{sosa00001@stud.uni-saarland.de} \\\And
  Rachel Bawden \\
   Inria, Paris, France \\ \\
  \texttt{rachel.bawden@inria.fr} \\}

\begin{document}
\maketitle
\begin{abstract}
Multilingual language models have shown impressive cross-lingual transfer ability across a diverse set of languages and tasks. 
To improve the cross-lingual ability of these models, some strategies include transliteration and finer-grained segmentation into characters as opposed to subwords. In this work, we investigate lexical sharing in multilingual machine translation (MT) from Hindi, Gujarati, Nepali into English. We explore the trade-offs that exist in translation performance between data sampling and vocabulary size, and we explore whether transliteration is useful in encouraging \textit{cross-script} generalisation. We also verify how the different settings generalise to unseen languages (Marathi and Bengali). 
We find that transliteration does not give pronounced improvements and our analysis suggests that our multilingual MT models trained on original scripts seem to already be robust to cross-script differences even for relatively low-resource languages.
Our code will be made publicly available.\footnote{\url{https://github.com/sonalsannigrahi/Multilingual_Strategy}}

\end{abstract}

\section{Introduction}

As research in natural language processing (NLP) moves towards handling an increasing number of languages \cite{aharoni-etal-2019-massively,fan2021beyond}, one of the key challenges is targeting low-resource and morphologically rich languages \cite{johnson-etal-2017-googles,magueresse2020low}. Multilingual language models such as mBERT \cite{devlin-etal-2019-bert} and XLM-R \cite{conneau-etal-2020-unsupervised} have shown surprising cross-lingual ability in zero and few-shot scenarios for a diverse set of languages \cite{wu-dredze-2020-languages}.

In order for low-resource languages to optimally benefit from data available for related and higher-resource languages, one research direction has been to explore what encourages better cross-lingual sharing in multilingual models, particularly in models that have joint vocabularies \cite{ha-etal-2016-toward,johnson-etal-2017-googles,aharoni-etal-2019-massively}.

%

One strategy for doing this is to preprocess the texts to reduce variation linked to differences in script and orthographic conventions, for example phonetisation, transliteration and transcription, in order to increase lexical overlap across languages. These pre-processing steps  have been used in the literature across several multilingual NLP tasks \cite{nakov-tiedemann-2012-combining,nguyen-chiang-2017-transfer,chakravarthi2019comparison,goyal-etal-2020-efficient,sun-etal-2022-alternative,muller-etal-2021-unseen,alabi-etal-2022-inria-almanach}. However, there is still some debate over how much transliteration helps in multilingual setups, despite it theoretically encouraging better lexical overlap, particularly for low-resource languages. For example, \newcite{pires-etal-2019-multilingual} found that transfer may be helped by increased lexical overlap (although it also works without it) and \newcite{K_Karthikeyan2020-sm} argue that lexical overlap has a negligible impact on transfer. \newcite{chakravarthi2019comparison} and \newcite{muller-etal-2021-unseen} found gains when transliterating, whereas for \newcite{alabi-etal-2022-inria-almanach}, results were less clear.

\begin{figure}[!t]
    \centering\small
    \includegraphics[width=0.4\textwidth]{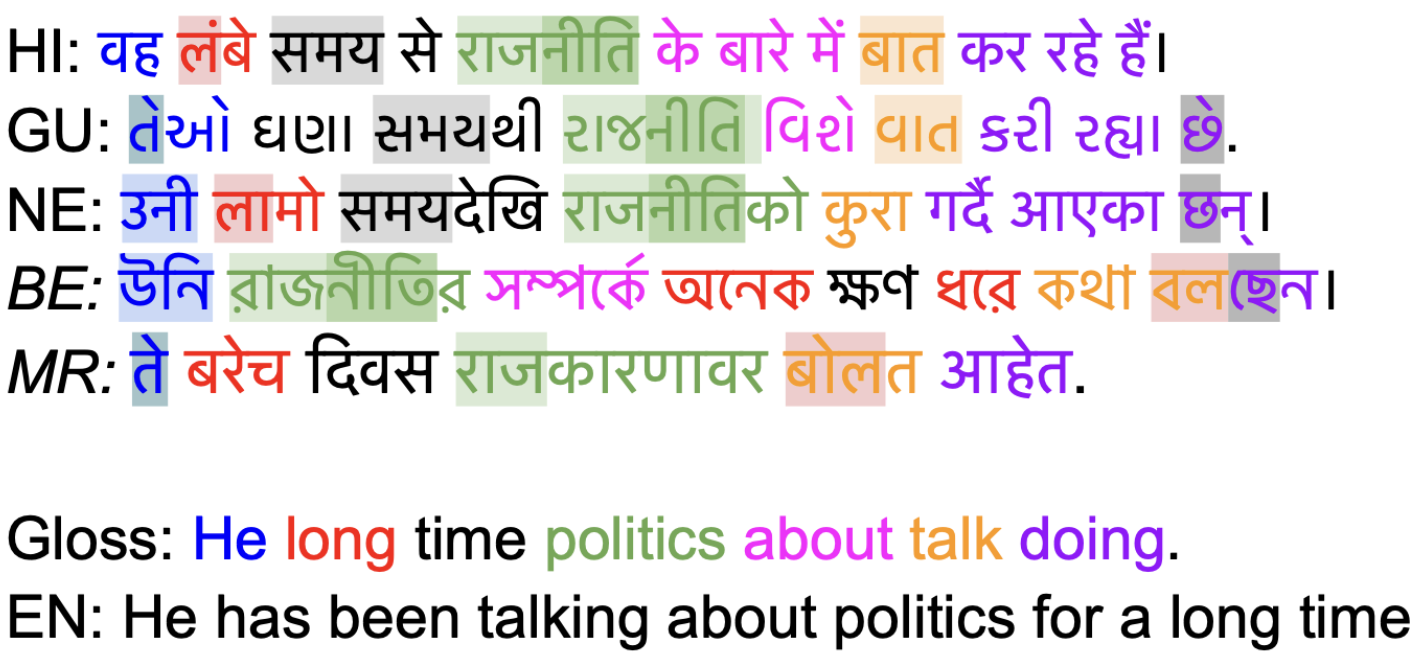}
    \caption{Illustration of partial lexical overlap in different scripts and languages (Hindi, Gujarati, Nepali, Bengali, Marathi). Highlighted text is an exact phonetic match at word or partial word coverage level. }
    \label{fig:lex}
\end{figure}

In this study, we build on this previous work to further investigate how lexical overlap can help multilingual machine translation (MT) by taking as a case study several Indian languages.
Figure~\ref{fig:lex} illustrates the degree of lexical overlap in the chosen languages of study: Hindi, Gujarati, Nepali, Bengali, and Marathi. Despite script differences, this example shows a sizeable amount of shared token overlap in terms of both characters and words. 

Focusing on the translation of these languages (Hindi, Gujarati, Nepali) into English, we explore the ideal parameter settings for multilingual MT (sampling vs.~segmentation size) and look at how transliterating into a single script (i.e.~Gujarati into Devanagari) may help performance. In addition, we look at how the trained models can transfer to other related languages (Bengali and Marathi) in zero- and few-shot settings.
%
We find that transliteration does not significantly help performance in our multilingual MT setup, even for the lowest-resourced language directions. Our analysis suggests that even with relatively little data, the multilingual model trained on the original scripts seems to learn a sufficient mapping between original and transliterated tokens, possibly making transliteration redundant. Even in zero- and few-shot transfer settings, we find only marginal improvements in the languages considered by using the multilingual model that uses transliteration as opposed to the multilingual model with the original scripts. 




\section{Related Work}


Multilingual models have been proposed for MT as well as other NLP tasks \cite{doddapaneni2021primer}. Within multilingual models, the promotion of lexical sharing has been the primary motivation to train multilingual models, which can especially aid low-resource languages \cite{conneau-etal-2020-unsupervised}. 

The choice of input unit has received a lot of attention, from the use of joint multilingual vocabularies \cite{sennrich-etal-2016-edinburgh,ha-etal-2016-toward,johnson-etal-2017-googles,aharoni-etal-2019-massively} and subword segmentation strategies \cite{sennrich-etal-2016-neural,kudo-richardson-2018-sentencepiece} to character-based \cite{kreutzer-sokolov-2018-learning} and byte-based \cite{xue-etal-2022-byt5} models. 
Other works have explored phonetisation \cite{liu-etal-2019-robust,rosales-nunez-etal-2019-phonetic} and transliteration/transcription in order to create a higher degree of lexical overlap in related languages that do not shared scripts \cite{nakov-tiedemann-2012-combining,nguyen-chiang-2017-transfer,chakravarthi2019comparison,goyal-etal-2020-efficient,muller-etal-2021-unseen,alabi-etal-2022-inria-almanach}. 


Cross-lingual word embedding spaces have been of interest as well. \newcite{chronopoulou-etal-2021-improving} map separately learnt embeddings to the same space, and other related works attempt to jointly learn  a shared embedding space for multiple languages. Cross-lingual transfer studies on multilingual models such as mBERT \cite{devlin-etal-2019-bert} have also shown the utility of multilingual pre-training especially for zero-shot transfer \cite{pires-etal-2019-multilingual}. They show that overlap can lead to better zero-shot transfer, although there can still be transfer with no overlap, as also seen by \newcite{k-etal-2021-analyzing}. \newcite{wu-dredze-2020-languages} also see a positive  correlation between lexical overlap and the zero-shot transfer performance. Additionally, \cite{oladipo2022an} experiment with effect of shared vocabulary spaces in multilingual setups for several low-resource African languages (Amharic, Hausa, and Swahili) and find that the number of languages used during pre-training has a positive effect on cross-lingual transfer only up to a certain point- which is improved by simply using a monolingual model with a multilingual tokeniser. 




Variation in data availability, scripts, and morpho-syntactic properties make adapting multilingual models to unseen languages challenging. Transliteration, which directly encourages lexical overlap, has shown positive results for texts in different scripts \cite{muller-etal-2021-unseen,chakravarthi2019comparison}. \newcite{muller-etal-2021-unseen} show that script plays a crucial role in improving transferability of multilingual models for languages that otherwise lag behind in performance. 
However, \newcite{alabi-etal-2022-inria-almanach} find that transcription (for Slavic languages) degraded rather than aided performance, with the hypothesis that the high-resource setup made transcription unnecessary, especially given the noise introduced by transcription.  
In our work, we study the role of transliteration in the case of multilingual MT for a set of lower-resource language directions, using related Indian languages with script differences. 



\section{Background on the Languages of Study}

Hindi, Nepali, Gujarati, Bengali, and Marathi are all Indo-Aryan languages, a sub-branch of the Indo-European language family, with speakers primarily concentrated in the Indian subcontinent. Hindi (excluding Urdu)\footnote{We exclude Urdu in the speaker counts, since Hindi and Urdu, although nearly identical phonetically, are written in different scripts (Devanagari and Arabic script respectively). This is an important distinction given that we focus on transliteration.} is spoken by approximately 340M L1 speakers (and 600M L1 or L2 speakers) and is considered to be the largest in terms of L1 speakers, whereas Nepali, Gujarati, Bengali, and Marathi have 16M, 57M, 272M, and 99M L1 speakers respectively.\footnote{Figures from Ethnologue, \url{https://www.ethnologue.com/insights/ethnologue200/}.} Hindi, Nepali, and Marathi share the same script (Devanagari) and also certain morpho-syntactic properties such as split ergativity and Subject-Object-Verb word order with constraint-based reordering allowed. Gujarati and Bengali each use their own scripts, although they are still considered closely related to the other Indo-Aryan languages, with both lexical and grammatical similarities. In particular, in both languages there exist many words that are an exact phonetic match with Hindi due to direct borrowing from Sanskrit. Due to these properties and the fact that the writing systems correspond well to the phonetic systems, transliteration from either the Gujarati and Bengali script into Devanagari is mostly straightforward (see Figure~\ref{fig:phonetic_overlap} for an example). 

\begin{table*}[!ht]

    \centering\small
    \begin{tabular}{lp{6.9cm}llrrr}
    \toprule
    & \multicolumn{3}{c}{Data sources} & \multicolumn{3}{c}{\#sentences} \\
     & Train & Dev & Test & Train & Dev & Test\\
    \midrule
    hi--en & Wikititles, HindEnCorp, IITB\footnotemark{} & WMT-dev14 & WMT-test14 & 1.3M & 520 & 2,507 \\
    ne--en & Bible,\footnotemark{} Ted2020,\footnotemark{} QED, GlobalVoices, GNOME, KDE  & Flores-dev & Flores-devtest & 115k & 997 & 1,012 \\
    gu--en & Bible, Wiki, Wikititles, Govin-clean, localisation & WMT-dev19 & WMT-test19 & 70k & 997 & 1,012\\
    mr--en & Bible-UEDIN, cvit-pib, jw, PMI, Ted2020, Wikimatrix & Flores-dev & Flores-devtest & 330k & 997 & 1,012\\
    be--en & alt, cvit-pib, jw, OpenSubtitles, PMI, Tanzil, Ted2020, Wikimatrix  & Flores-dev & Flores-devtest & 86k & 997 & 1,012 \\
    \bottomrule
    \end{tabular}
    \caption{Data sources and dataset sizes for each language pair.}
    \label{tab:data:stats-sources}
\end{table*}

\addtocounter{footnote}{-2}
\footnotetext{\cite{kunchukuttan-etal-2018-iit}}
\addtocounter{footnote}{1}
\footnotetext{\cite{christodouloupoulos2015massively}}
\footnotetext{\cite{reimers-2020-multilingual-sentence-bert}}
\section{Experiments}

We study the effect of transliteration for multilingual MT to test the hypothesis that increased lexical overlap between the training languages could boost performance, particularly for lower-resourced language pairs.
We  study two different scenarios: (i)~an \textit{in-language} scenario, whereby we train and evaluate on the same set of language pairs, namely Hindi (hi), Nepali (ne), and Gujarati (gu) into English, and (ii)~zero- and few-shot transfer (via fine-tuning) of these models to two unseen related language pairs, namely Marathi (mr) and Bengali (bn) into English. We compare models trained on the original scripts and after transliteration (i.e.~Gujarati is transliterated into Devanagari).

Since the aim of transliteration is to increase lexical overlap between the languages, 
we make sure to monitor for the degree of tokenisation, as well as data sampling, both crucial parameters in multilingual MT performance that directly affect token overlap, to ensure a fair comparison. 

\subsection{Data}\label{sec:data}

The chosen languages cover a variety of scripts (Devanagari, Gujarati, and Bengali) as illustrated in Figure~\ref{fig:lex}. Table~\ref{tab:data:stats-sources} lists the data sources and sizes used 
(ranging from 65k sentences for gu--en to 1M sentences for hi--en after post-processing). 

We clean the data by normalising punctuation, and removing duplicate sentence pairs from the training data. For experiments involving transliteration, we use the IndicNLP toolkit\footnote{\url{https://github.com/anoopkunchukuttan/ indic_nlp_library}} \cite{kunchukuttan2020indicnlp}  to transliterate Gujarati and Bengali scripts into the Devanagari script. For subword segmentation, we use the Sentencepiece toolkit \cite{kudo-richardson-2018-sentencepiece} and the BPE strategy \cite{sennrich-etal-2016-neural} to train joint models covering the specific training languages for each model, i.e.~the source and target language for bilingual models and Hindi, Gujarati, Nepali and English for the multilingual ones. We test a range of vocabulary sizes: 4k, 8k, 16k and 32k for the multilingual models and 4k, 8k, 10k for the bilingual models.\footnote{Preliminary experiments showed that larger vocabulary sizes degraded the performance.} 

\begin{figure}
    \centering
    \includegraphics[width=0.38\textwidth]{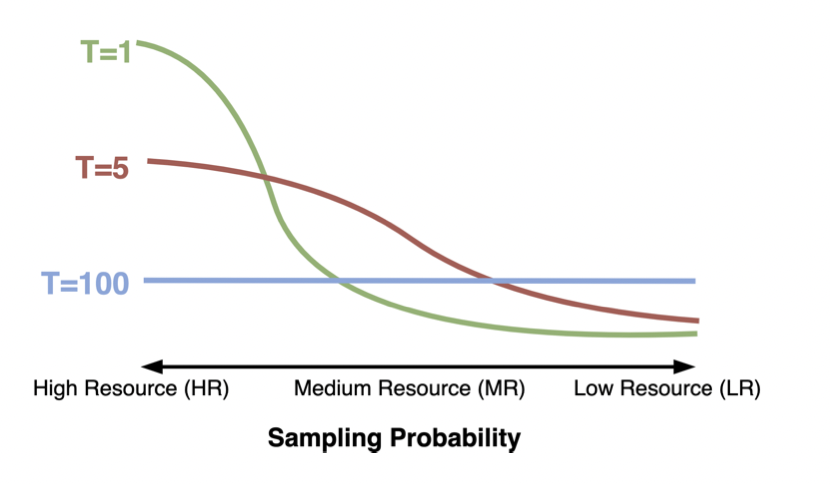}
    \caption{Illustration of data distribution with temperature sampling, taken from \cite{arivazhagan2019massively}.}
    \label{fig:tt_sample}
\end{figure}

Due to differences in the amount of data available, we use temperature sampling to address imbalances  \cite{fan2021beyond}. We sample data with probability $p_l$ from each language pair, $l$ with $D_l$ size parallel corpora, included in the data during training of the SentencePiece models and the training of the multilingual MT model as follows:
\[
p_{l} \propto (\frac{D_{l}}{\sum_{k}D_{k}})^{\frac{1}{T}},
\]
where $T$ corresponds to the temperature, which adjusts how much the original distribution is favoured ($T$=1) versus a more uniform distribution of the data (higher $T$ value)  as illustrated in Figure \ref{fig:tt_sample}.

%
We test the temperature values 1.2, 1.5 and 1.8.\footnote{Preliminary experiments showed that more extreme (higher) values worked less well, despite these being used previously in the literature \cite{aharoni-etal-2019-massively}.}


\subsection{Models}


We train multilingual models for Hindi, Gujarati, and Nepali into English for the vocabulary sizes and temperatures specified in Section~\ref{sec:data}, comparing models using (i)~the original scripts and (ii)~when Gujarati is transliterated into Devanagari (i.e.~all sources languages use Devanagari). We compare these models to bilingual baselines for each of the three main language pairs, trained in the same way but only with the source and target languages concerned. 

All models are transformers as implemented in Fairseq \cite{ott2019fairseq}. We use the following default parameters unless 
stated otherwise:\footnote{\url{https://github.com/facebookresearch/fairseq}} 6 encoder and decoder layers with 512 embedding dimension, 2048 FFN embedding dimension, and 8 heads for both the encoder and decoder. For the multilingual models, we use a shared encoder to promote language sharing.
All models are trained using the Adam optimiser with a learning rate of $3e{-5}$. All the models, multilingual and bilingual, use the same hyperparameters. Models are trained until convergence and the best model is selected according to the BLEU score on the development set. 
We evaluate using BLEU \cite{papineni-etal-2002-bleu} using the SacreBLEU toolkit \cite{post-2018-call}.\footnote{Signature =\texttt{nrefs:1|case:mixed|eff:no|tok:13a
 smooth:exp|version:2.0.0}}


\section{Results}
\begin{table*}[!ht]
\begin{subtable}[h]{0.3\textwidth}
\centering\small
\setlength\tabcolsep{4pt}
\begin{tabular}{lrrr}
    \toprule
    Vcb. & gu$\rightarrow$en & hi$\rightarrow$en & ne$\rightarrow$en \\
        \midrule
        &  \multicolumn{3}{c}{\textit{Original}}\\
        \midrule
    4k & 3.87 & 10.12 & 2.06  \\
    8k & 3.95 & 10.44 & 2.33 \\
    10k & 4.12 & 12.32 & 2.37  \\
    \midrule
    &  \multicolumn{3}{c}{\textit{Transliterated}}\\
    \midrule
    4k & 3.48 & -- & -- \\
    8k & 3.68 & -- & -- \\
    10k & 4.11 & -- & -- \\
    \bottomrule
  \end{tabular}
  \caption{Bilingual models. 
  }
  \label{tab:bilingual-baselines2}
\end{subtable}
\begin{subtable}[h]{0.65\textwidth}
\setlength\tabcolsep{4pt}
\centering\small
\begin{tabular}{lrrrrrrrrrrrr}
    \toprule
    
    &
      \multicolumn{3}{c}{gu$\rightarrow$en} &
      \multicolumn{3}{c}{hi$\rightarrow$en} &
      \multicolumn{3}{c}{ne$\rightarrow$en} \\
      Temp. & 1.2 & {1.5} & {1.8} & {1.2} & {1.5} & {1.8} & {1.2} & {1.5} & {1.8}  \\
      \midrule
        Vcb. $\downarrow$  &  \multicolumn{9}{c}{\textit{Original}}\\
           \midrule
    char & 11.30 & 11.45 & 11.63 & 14.78 & 15.12 & 15.64 & 11.02 & 10.46 & 10.89 \\
    4K & 11.10 & 11.40 & \textbf{\textit{11.82}} & 15.03 & 14.14 & 14.34 & 11.12 & \textbf{12.10} & \textbf{\textit{12.52}}\\
    8K & \textbf{11.46} & \textbf{11.69} & 11.58  & 15.01 & 14.60 & 14.66 & \textbf{11.85} & 11.80 & 11.79 \\
    16K & 11.42 & 9.99 & 11.59 & 15.11 & 14.70  & \textbf{14.78}& 11.73 & 10.44 & 11.56 \\
    32K & 11.37 & 11.11 & 11.01& \textbf{\textit{15.32}} & \textbf{14.76} & 14.57 & 11.60 & 11.20 & 11.31\\
    \midrule
     & \multicolumn{9}{c}{\textit{Transliterated}}\\
     \midrule
    char & \textbf{11.67} & \textbf{11.82} & \textit{\textbf{11.96}} & 12.78 & 13.35 & 13.41 & 10.87 & 11.21 & 11.30 \\
    4K & 11.42 & 11.65 & 11.78 & 13.32 & 13.28 & 13.61 & \textbf{12.23} & \textbf{12.52} & \textbf{\textit{12.56}}\\
    8K &11.21 & 11.34 & 11.68  & 13.28 & 13.56 & 13.55 & 11.32 & 11.50 & 11.87 \\
    16K & 11.12 & 11.46 & 11.54 &\textbf{ 13.10} & \textbf{\textit{14.38}} & \textbf{14.33} & 11.11 & 11.24 & 11.73 \\
    32K & 11.00 & 11.08 & 11.56 & 13.14 & 13.44 & 13.75 & 11.10 & 11.20 & 11.65\\
    
    \bottomrule
  \end{tabular}
  \caption{\label{tab:multilingual-regular}Multilingual models. 
  }
  \end{subtable}
  \caption{\label{tab:results-all}BLEU scores for bilingual baseline and multilingual models (original and transliterated) for different vocab sizes (Vcb.) and temperature values (for multilingual models only) averaged over three runs with different starting seeds. Bold represent the best score for each temperature, italics represents best score overall.}
\end{table*}

The main results are shown in Table~\ref{tab:bilingual-baselines2} for bilingual models and Table~\ref{tab:multilingual-regular} for multilingual models.

\subsection{Does multilinguality help?}\label{sec:results1}

We start by evaluating whether multilinguality helps by comparing the models trained on original scripts.
Tables~\ref{tab:bilingual-baselines2} and~\ref{tab:multilingual-regular} summarise these results for each of the language directions considered (hi$\rightarrow$en, gu$\rightarrow$en, ne$\rightarrow$en). For the lower-resourced pairs, the bilingual MT models perform poorly (less than 5 BLEU points). However, these scores are greatly improved in the multilingual MT model 
(ne$\rightarrow$en and gu$\rightarrow$en achieve 12.52 and 11.82 BLEU respectively as the highest scores across all configurations tested). 
This performance jump demonstrates the large gains that can be observed via knowledge transfer in multilingual models, confirming previous work \cite{dabre2020survey}.


In terms of temperature and vocabulary size, our multilingual results are coherent with the existing literature \cite{cherry-etal-2018-revisiting,kreutzer-sokolov-2018-learning}, which suggests that using smaller sub-word tokens perform better in low-resource settings due to their improved ability to generalise;  
for the lower-resource language pairs (\{ne,gu\}$\rightarrow$en) a higher temperature and smaller vocabulary size combination was preferred,\footnote{4k vocabulary size, $T$=1.8.} while for the higher-resource language pair (hi$\rightarrow$en) a lower temperature and larger vocabulary size combination was better.\footnote{32k vocabulary size, $T$=1.2.} 

\subsection{Is Transliteration Useful?}\label{sec:translit}

Our hypothesis was that by transliterating Gujarati into the Devanagari script, we might be able to see gains through increased lexical sharing amongst the three source languages in a multilingual setup.

As a control experiment to test the impact of transliteration outside of the multilingual setup, we compare results for the bilingual model using the original Gujarati script and when transliterated into Devanagari script (Table~\ref{tab:bilingual-baselines2} \textit{Transliterated}). The transliterated model performs slightly worse than the original bilingual model (0.24\% decrease between the highest scores) suggesting that transliteration may be introducing ambiguity or noise, as also suggested by \newcite{alabi-etal-2022-inria-almanach}.  
For the multilingual models (Table~\ref{tab:multilingual-regular}), in the case of hi$\rightarrow$en (the highest-resourced language) transliteration leads to a 8.6\% decrease in the BLEU score. 
This decrease does not appear for gu$\rightarrow$en and ne$\rightarrow$en, where instead marginal improvements of 0.08 and 0.04 BLEU between the highest scores respectively are observed. However this improvement is not as large as suggested by some previous work \cite{muller-etal-2021-unseen}. The results here could suggest that the original model might be sufficiently capturing the same level of information regarding token overlap as transliteration.

Overall compared to the original model in both the bilingual and multilingual setup, we find the improvements from transliteration (when applicable) to be not as pronounced.



\subsection{Mapping Tokens in the Multilingual Embedding Space}

The lack of significant improvement in in-language performance for the transliterated model is in line with results seen by \newcite{alabi-etal-2022-inria-almanach}, but is more surprising given that we test on two lower-resourced language pairs. 
So does this mean that the original model is already able to map between tokens written in different scripts?

\begin{figure}
    \centering
    \includegraphics[width=0.35\linewidth]{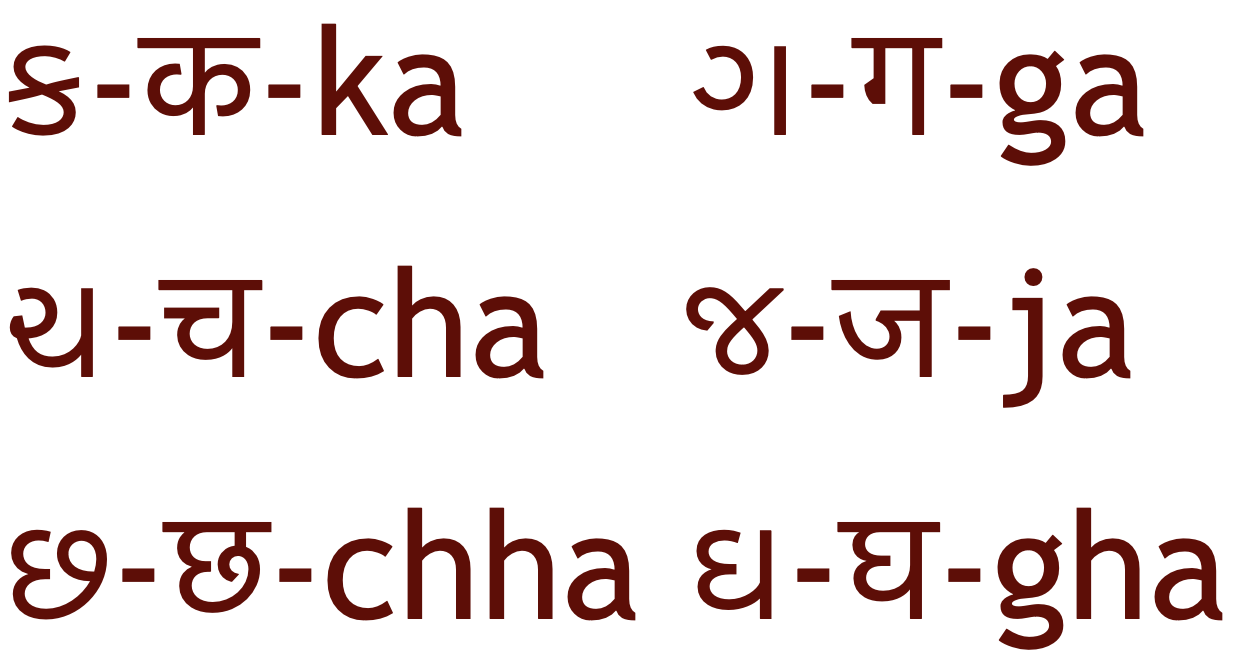}
    \caption{Examples of six consonants and their realisation in Gujarati, Devanagari and Latin scripts.}
    \label{fig:phonetic_overlap}
\end{figure}

\begin{figure}[ht]
    \centering
    \includegraphics[width=0.33\textwidth]{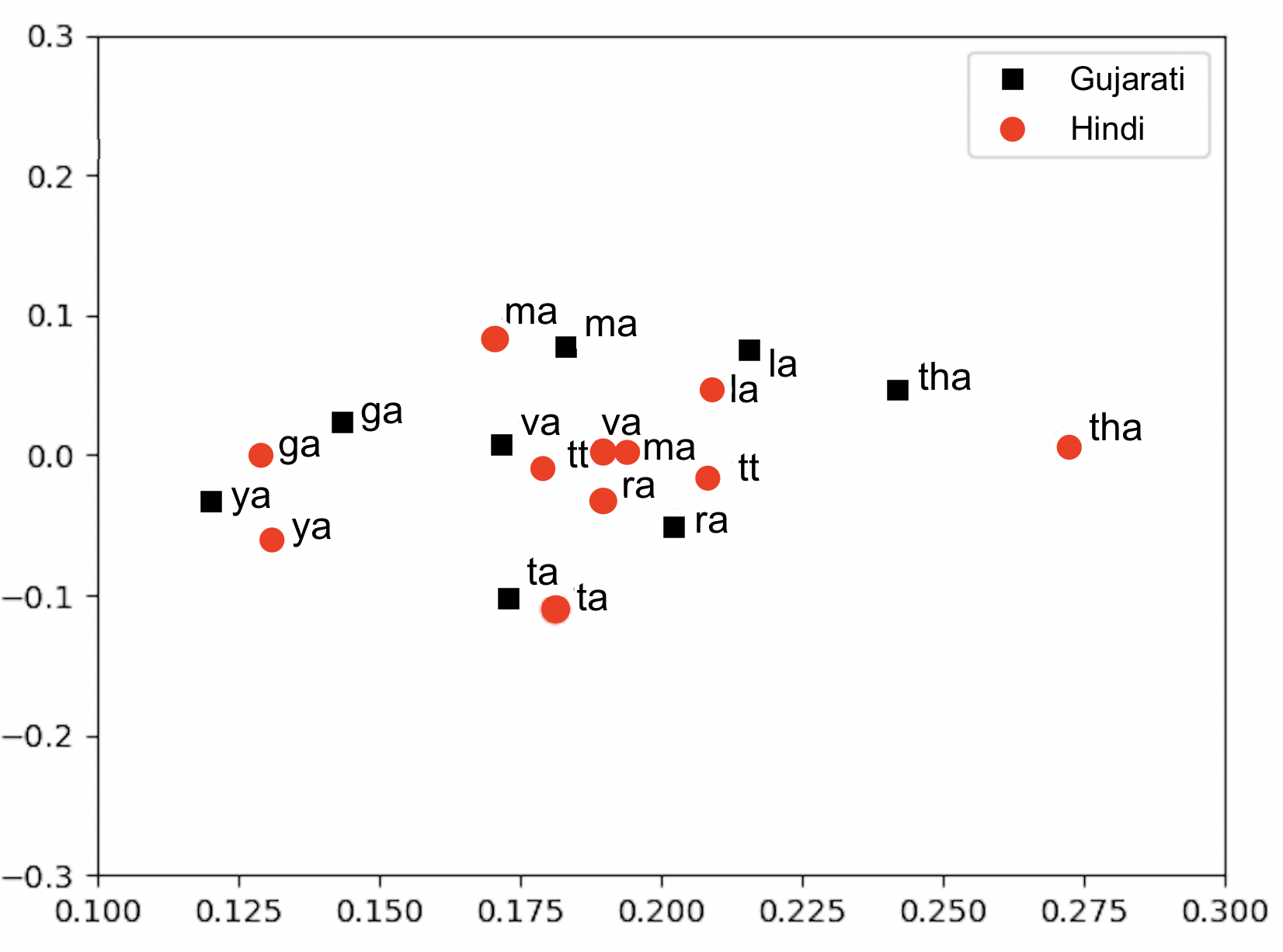}
    \caption{PCA projection of the multilingual embedding space (Original, 16k vocabulary size, T=1.5), where labelled points are a selection of phonetically equivalent tokens in Devanagari script (red dots) and Gujarati script (black squares).}
    \label{fig:emb_proj}
\end{figure}

To test this, we look at the similarity of tokens that are phonetically equivalent aside from being written in different scripts. Figure~\ref{fig:phonetic_overlap} shows some examples of Gujarati and Devanagari characters and (for illustration purposes) their romanised phonetic equivalents. 
Figure~\ref{fig:emb_proj} illustrates the embedding projection of the original multilingual model (16k vocab size, $T$=1). We use PCA to perform dimension-reduction, and we use 10000 tokens from the vocabulary to learn the embedding space. We observe that phonetically equivalent tokens in the Devanagari and Gujarati scripts are mapped reasonably close together in this embedding space suggesting that despite script differences, the model seems to have learnt similar representations.  

\begin{table}[!ht]
    \centering\small
    \begin{tabular}{lrrrrrr}
    \toprule
     & \multicolumn{6}{c}{Hindi} \\
     Gujarati $\downarrow$ & Pa & Ma & Da & Ka & Fa & Avg.\\
    \midrule
     Pa & \textbf{0.73} & 0.12 & 0.02 & 0.14 & 0.02 & 0.01\\
     Ma & 0.18 & \textbf{0.75} & 0.05 & 0.20 & 0.13 & 0.04 \\
     Da & 0.02 & 0.25 & \textbf{0.35} & 0.26 & 0.03 & 0.02 \\
     Ka & 0.15 & 0.26 & 0.02 & \textbf{0.66} & 0.01 & 0.03 \\
     Fa & 0.02 & 0.25 & 0.12 & 0.20 & \textbf{0.45} & 0.01\\
     Avg. & 0.01 & 0.02 & 0.02 & 0.01 & 0.03 & - \\
    \bottomrule
    \end{tabular}
    \caption{Cosine similarity scores between phonetically identical units in Devanagari (horizontal) and Gujarati (vertical) scripts with an average score (Avg.) between all other tokens.} 
    \label{tab:cosine_sim}
\end{table}




\subsection{Cross-script Robustness}

\begin{figure}[!ht!]
    \centering
    \includegraphics[width=0.3\textwidth]{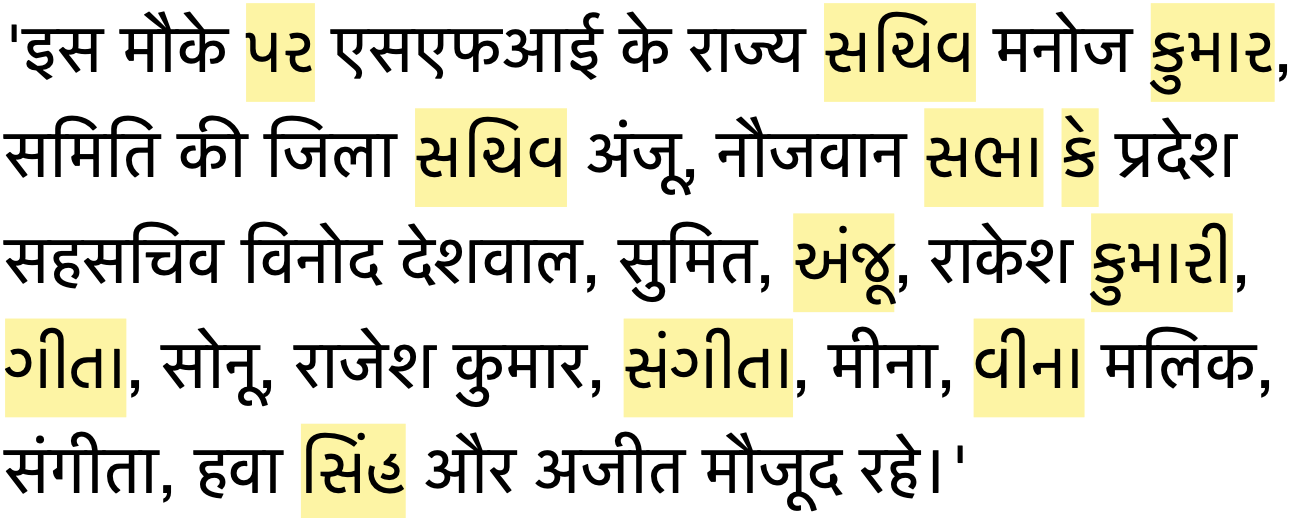}
    \caption{Example of Hindi text in Devanagari script with 30\% of words  transliterated into Gujarati script (highlighted). }
    \label{fig:xscript_ex}
\end{figure}

\begin{figure}[!ht!]
    \centering
    \includegraphics[width=0.4\textwidth]{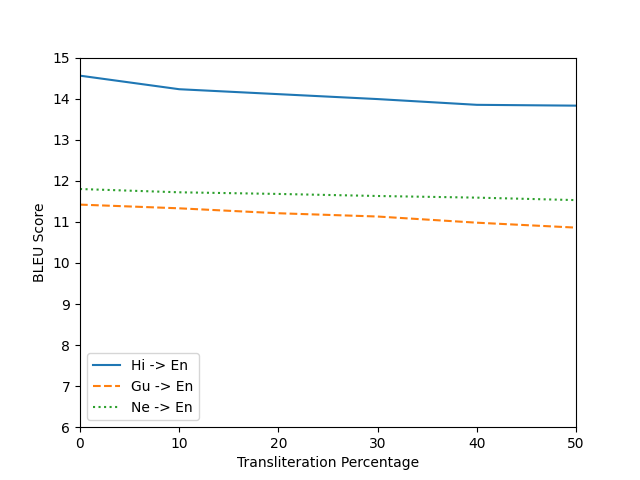}
    \caption{BLEU scores of the multilingual model (8k vocab, T=1.5) with an increasing percentage of cross-script switching.}
    \label{fig:incremental}
\end{figure}

We additionally experiment with cross-script switching to test how robust the original multilingual model is to changes in the script being used, as it appears to provide reasonably similar mappings between the same tokens written in different scripts. We artificially create texts with increasing percentages of transliteration into a different script seen by the model and evaluate the model at inference on these texts in a zero-shot fashion. 
For Devanagari text (in Hindi and Nepali), we transliterate parts of the text into Gujarati and vice versa. We randomly select a certain percentage of words to transliterate in each sentence. Figure~\ref{fig:xscript_ex} shows an example of cross-script switching for Hindi with 30\% of words transliterated into Gujarati. We plot the BLEU scores of the different model configurations against the percentage of word-level transliteration in the test set in Figure~\ref{fig:incremental}. For brevity, we only plot results with $T=1.5$ and subword vocabulary size of 16k tokens in the original multilingual model that keeps the scripts as they are.\footnote{We observe similar results across the other temperature-vocabulary size configurations.} 

Although there is a downward trend in the BLEU scores, there is no significant degradation in performance with increasingly transliterated texts (only -0.2 BLEU with 50\% transliteration for gu$\rightarrow$en). The degradation of performance in the case of Hindi is more pronounced (-0.7 BLEU with 50\% transliteration for hi$\rightarrow$en). It is to be noted that in the earlier experiments (Table \ref{tab:multilingual-regular}) we found similar performance drops in Hindi between the original multilingual model and the transliterated multilingual model. This suggests that transliteration may not be a particularly useful strategy to promote lexical sharing as the models appear to already be reasonably robust to script differences.

\subsection{How Well do Models Generalise to Unseen Languages?} 

\begin{figure*}
     \centering
     \begin{subfigure}[b]{0.38\textwidth}
         \centering
         \includegraphics[width=\textwidth]{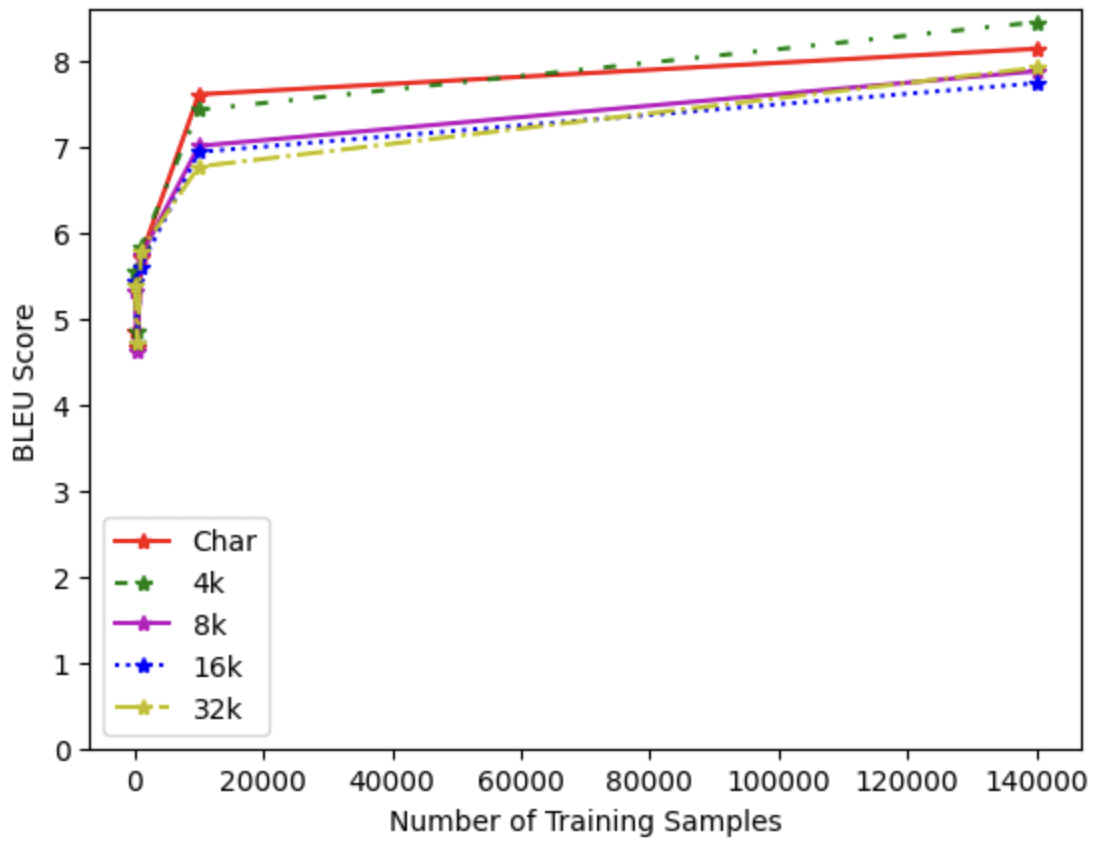}
         \caption{mr$\rightarrow$en with the original model}
         \label{fig:mr_og}
     \end{subfigure}
     \begin{subfigure}[b]{0.38\textwidth}
         \centering
         \includegraphics[width=\textwidth]{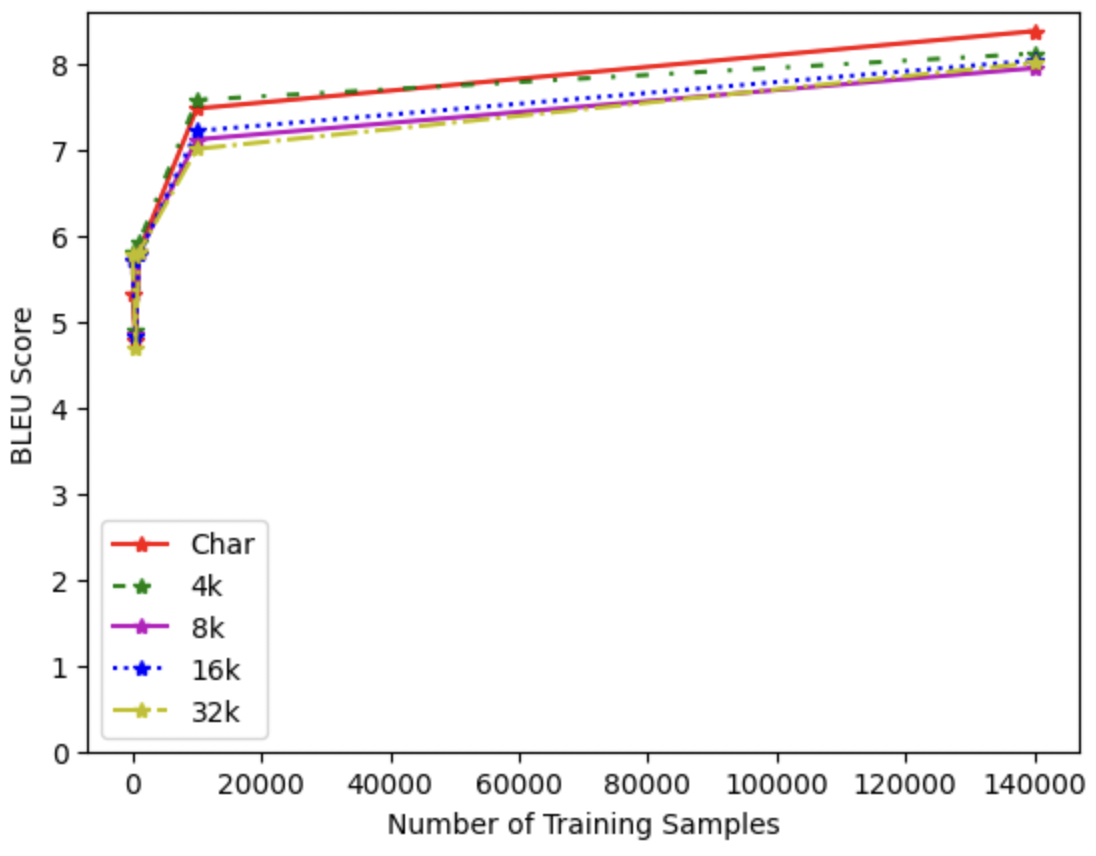}
         \caption{mr$\rightarrow$en with the transliterated model}
         \label{fig:mr_tl}
     \end{subfigure}
     \begin{subfigure}[b]{0.38\textwidth}
         \centering
         \includegraphics[width=\textwidth]{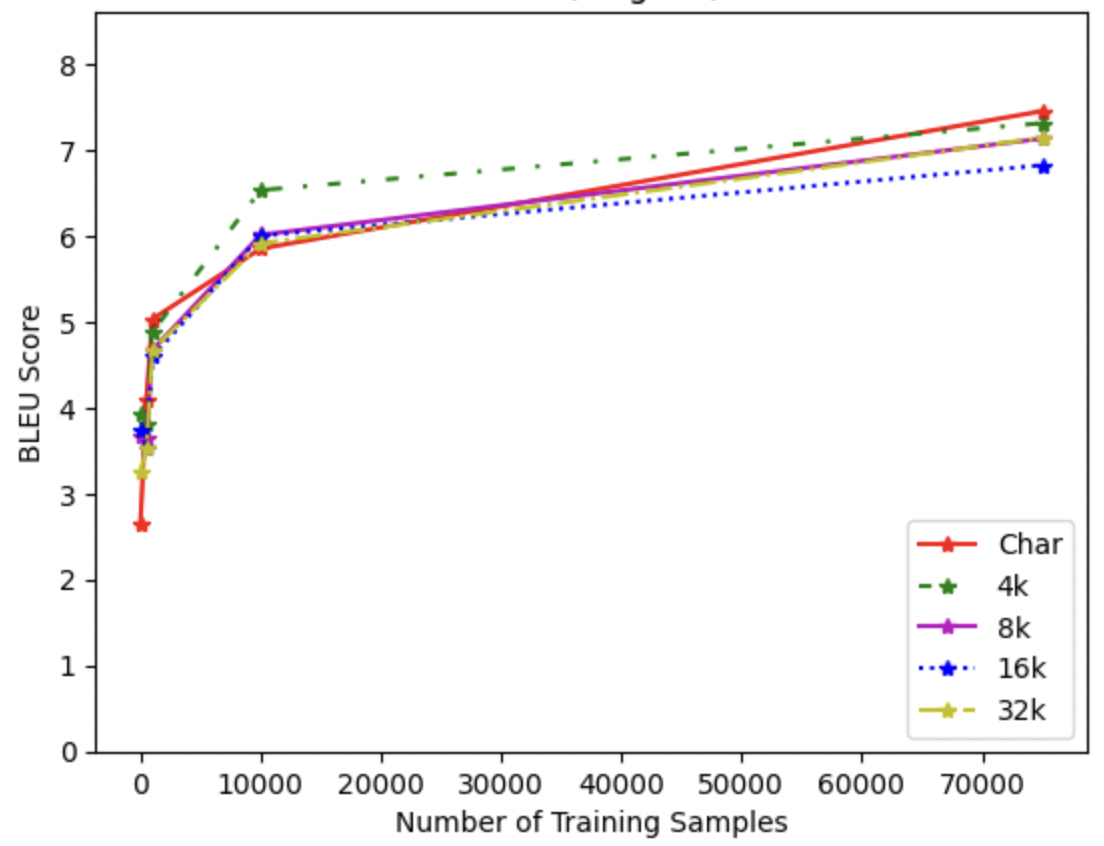}
         \caption{bn$\rightarrow$en with the original model}
         \label{fig:be_og}
     \end{subfigure}
     \begin{subfigure}[b]{0.38\textwidth}
         \centering
         \includegraphics[width=\textwidth]{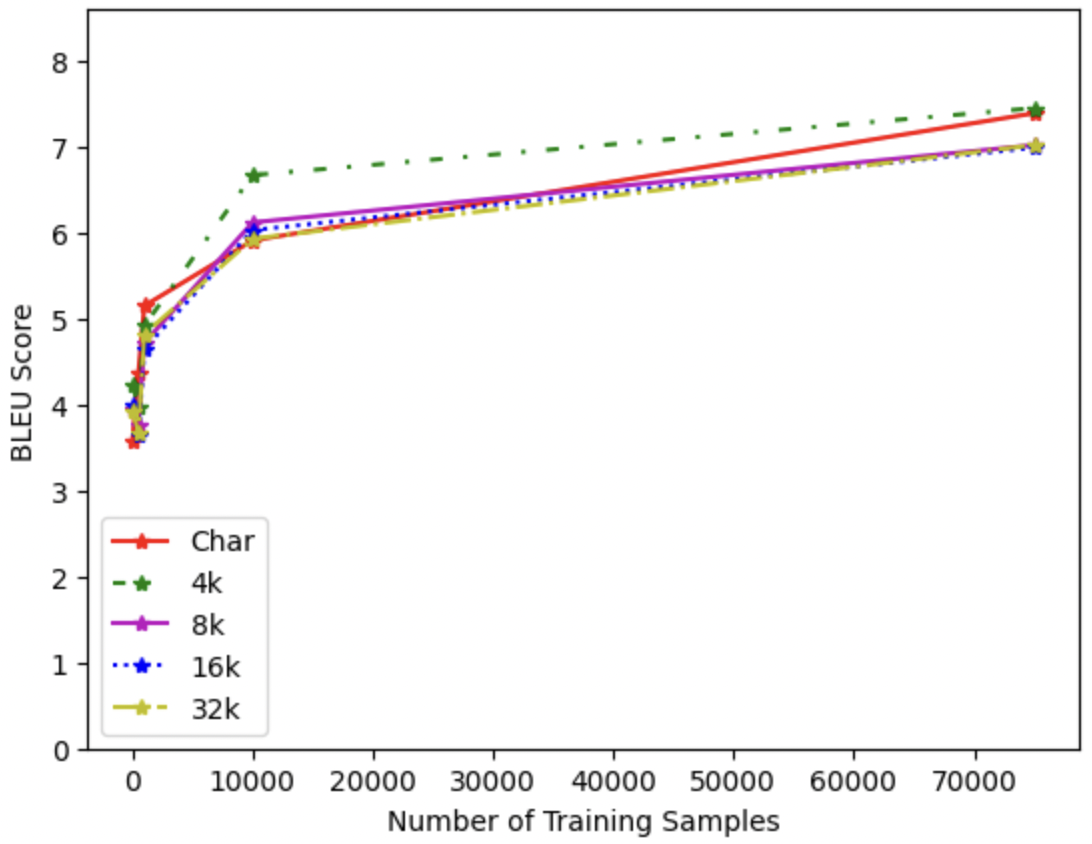}
         \caption{bn$\rightarrow$en with the transliterated model}
         \label{fig:be_tl}
     \end{subfigure}
        \caption{BLEU scores after fine-tuning on different amounts of supervised training data (starting with zero-shot performance, i.e.~no language-pair-specific data) for both the original multilingual model and after transliteration with varous across mvocabulary sizes: char, 4k, 8k, 16k and 32k. Only the best performing temperature value is plotted for clarity and space reasons. }
        \label{fig:few_shot}
\end{figure*}

Lastly, we study the models' ability to generalise to previously unseen but related languages. \newcite{adelani-etal-2022-thousand} find that the most effective strategy for transferring to additional languages is to use a small quantities of high-quality data. In our case, we do not fine-tune a large pre-trained language model but rather a multilingual translation model trained on Hindi, Nepali, Gujarati, and English. We therefore expect gains to be more limited than those demonstrated in \cite{adelani-etal-2022-thousand}. 

We evaluate zero-shot and few-shot transfer from the multilingual models with and without transliteration into two languages that share morphological similarities with the previous languages: Marathi (written with the Devanagari Script) and Bengali (written with the Bengali Script).\footnote{Across all models (original and transliterated) we first transliterate Bengali into Devanagari script in order to use the learned representations of the model. We leave Marathi in its original script (Devanagari)} In this setup we incrementally increase the amount of data used to fine-tune different models (zero-shot and 500, 1k, and 10k samples for the few-shot settings). We also include a topline in which we finetune the same models on all the available data (140k sentence pairs for mr$\rightarrow$en and 75k sentence pairs for bn$\rightarrow$en). Figure~\ref{fig:few_shot} summarises our results. The raw results are in Appendix~\ref{sec:general}. 


The results of the zero-shot performance of the configurations illustrated\footnote{We plot the best result for each vocabulary size in {\ char, 4k, 8k, 16k, 32k }\ } show that there is minimal generalisation of our multilingual model (original and transliterated) to new languages, despite their linguistic relatedness, with BLEU scores under 6 for both language directions. Using transliteration, the zero-shot transfer results are marginally improved (
an increase from 5.56 to 5.81 BLEU for mr$\rightarrow$en  
and from 3.93 to 4.23 BLEU for bn$\rightarrow$en when using the transliterated rather than original model). 

In the few-shot setup, similar to the results in Section~\ref{sec:results1} for the lower-resourced language pairs, smaller vocabulary sizes and higher temperature values are preferred ($T$=1.8 and either 5k or character-based segmentation). 
As with the zero-shot setup, marginal improvements with transliteration are observed in the few-shot setup. This result agrees with our earlier results (Section~\ref{sec:translit}), which show that transliteration does not provide significant gains, possibly as the original multilingual model is already robust to cross-script differences.

 

\section{Conclusions}

In this work, we studied language sharing in multilingual MT of several languages in the Indo-Aryan language family (Gujarati, Nepali, and Hindi into English). Experimenting with sampling temperature and vocabulary size, we compare multilingual models using the original scripts and when transliterating Gujarati into the same script as Nepali and Hindi (Devanagari). 
Surprisingly, even for the low-resource language directions (gu$\rightarrow$en and ne$\rightarrow$en), we find that transliteration is not particularly helpful. It seems that our multilingual models using the original scripts are able to correctly map phonetically equivalent tokens together, as suggested by (i)~our analysis of the embeddings of identical characters across scripts and (ii)~testing the robustness of the model to cross-script switching. Finally, we test how well the models transfer to unseen related languages (Marathi and Bengali into English). We find that the model with transliteration does not perform significantly better with respect to generalisation to unseen languages, further supporting our previous findings. 



\section{Acknowledgments}
R. Bawden's participation was partly funded by her chair position in the PRAIRIE institute, funded by the French national agency ANR as part of the ``Investissements d'avenir'' programme under the reference ANR-19-P3IA-0001. The work was also funded by R. Bawden's Emergence project, DadaNMT, funded by
Sorbonne Université.

\bibliography{anthology,custom}
\bibliographystyle{eamt23}
\clearpage
\appendix

\section{Generalisation of Models}\label{sec:general}

Table~\ref{tab:fewshot-results} reports results for the zero-shot and few-shot set-up for Marathi-English and Bengali-English. We use samples of sizes 500, 1k, 10k, and further report a fine-tuning topline, which uses all available data for each of the language pairs. Similar to the earlier setups, we evaluate vocabulary sizes in \{ character, 4k,8k,16k,32k \} and temperature values in \{ 1.2,1.5, 1.8 \}. 
\begin{table*}[!ht]
\setlength\tabcolsep{4pt}
\centering\small
\begin{tabular}{lrrrrrrrrrrrrrrr}
    \toprule
    & \multicolumn{15}{c}{\textbf{\#fine-tuning examples}} \\
    & \multicolumn{3}{c}{\textbf{0}} 
    & \multicolumn{3}{c}{\textbf{0.5k}} & \multicolumn{3}{c}{\textbf{1k}} & 
    \multicolumn{3}{c}{\textbf{10k}} &\multicolumn{3}{c}{\textbf{full set}}\\
        \midrule
   & 1.2 & 1.5 & 1.8  & 1.2 & 1.5 & 1.8 & 1.2 & 1.5 & 1.8 & 1.2 & 1.5 & 1.8 & 1.2 & 1.5 & 1.8\\
    \midrule
    & \multicolumn{15}{c}{\textit{Original}}\\
      \midrule
     & \multicolumn{15}{c}{\textbf{mr$\rightarrow$en}}\\ 
     \midrule
    char & 4.23 & 4.54 & \textbf{4.86} & 4.50 & 4.61 & \textbf{4.68} & 5.58 & 5.63 & \textbf{5.72}& 7.37 & 7.58 & \textbf{\textit{7.61}} & 8.07 & \textbf{8.14} & 8.02\\
    4K & 4.89 & 5.12 & \textbf{\textit{5.56}} & 4.12 & 4.72 & \textbf{\textit{4.86} }& 5.23 & 5.65 & \textbf{\textit{5.83}} & 6.98 & 7.12 & \textbf{7.43} & 8.02 & 8.34 & \textbf{\textit{8.45}} \\
    8K & 4.45 & 4.83 &\textbf{ 5.32} & 4.03 & 4.53 & \textbf{4.62} & 5.11 & 5.44 &\textbf{ 5.73} & 6.87 & 6.99 & \textbf{7.01} & 7.63 & 7.72 & \textbf{7.88} \\
    16K & 4.36 & 4.49 &\textbf{ 5.43} & 4.41 & 4.46 & \textbf{4.72} & 5.08 & 5.39 & \textbf{5.61 }& 6.76 & 6.87 & \textbf{6.94 }& 7.58 & 7.63 & \textbf{7.74}\\
    32K & 4.11 & 4.35 & \textbf{5.40 }& 4.52  & 4.68 & \textbf{4.71 }& 5.33 & 5.46 & \textbf{5.79} & 6.54 & 6.57 & \textbf{6.77 }& 7.41 & 7.64 & \textbf{7.92} \\
    \midrule
     & \multicolumn{15}{c}{\textbf{bn$\rightarrow$en}}\\ 
     \midrule
    char & 2.53 & 2.61 & \textbf{2.64} & 4.03 & 4.08 & \textbf{\textit{4.10}} & 4.98 & 5.01 & \textbf{\textit{5.03}} & 5.72 & 5.78 & \textbf{5.85 }& \textbf{\textit{7.45}} & 7.36 & 7.40\\
    4K & 3.31 & 3.41 & \textbf{\textit{3.93}} & 3.02 & 3.43 & \textbf{3.81} & 4.40 & 4.51 &\textbf{ 4.88} & 6.12 & 6.49 & \textbf{\textit{6.53}} & 6.98 & 7.08 & \textbf{7.31} \\
    8K & 3.50 & 3.55 & \textbf{3.67} & 3.01 & 3.48 & \textbf{3.65} & 4.35 & 4.48 & \textbf{4.67} & 5.56 & 5.93 & \textbf{6.01} & 6.48 & 6.75 & \textbf{7.13} \\
    16K & 3.65 & 3.70 & \textbf{3.74} & 3.00 & 3.49 & \textbf{3.52} & 4.28 & 4.37 & \textbf{4.59} & 5.71 & 5.83 &\textbf{ 6.00} & 6.16 & 6.37 & \textbf{6.82}\\
    32K & 3.21 & 3.25 &\textbf{ 3.26} & 3.07  & 3.35 & \textbf{3.52} & 4.36 & 4.48 & \textbf{4.67} & 5.74 & 5.86 & \textbf{5.91} & 6.80 &\textbf{ 7.14} & 7.04 \\
    \midrule
    & \multicolumn{15}{c}{\textit{Transliterated}}\\
      \midrule
     & \multicolumn{15}{c}{\textbf{mr$\rightarrow$en}}\\ 
     \midrule
    char & 5.02 & 5.12 & \textbf{5.33} & 4.66 & 4.76 & \textbf{4.78} & 5.73 & \textbf{5.81} & 5.80 & 7.32 & 7.46 & \textbf{7.48} & 8.20 & 8.34 & \textbf{\textit{8.38}}\\
    4K & 5.02 & 5.33 & \textbf{\textit{5.81}} & 4.51 & 4.73 & \textbf{\textit{4.91}} & 5.61 & 5.72 & \textbf{\textit{5.92}} & 7.11 & 7.34 & \textbf{\textit{7.58}} & 8.10 & \textbf{8.12} & 7.99 \\
    8K & 5.24 & 5.41 &\textbf{ 5.71} & 4.34 & 4.65 & \textbf{4.85 }& 5.50 & 5.61 & \textbf{5.80} & 6.95 & 7.02 & \textbf{7.12 }& 7.71 & 7.86 &\textbf{ 7.95} \\
    16K & 5.15 & 5.41 & \textbf{5.71 }& 4.22 & 4.60 & \textbf{4.83 }& 5.48 & 5.65 & \textbf{5.78} & 6.92 & 6.98 & \textbf{7.22} & 7.95 & 7.98 & \textbf{8.04}\\
    32K & 5.17 & 5.76 & \textbf{5.78 }& 4.40  & 4.70 & \textbf{4.70} & 5.45 & 5.58 & \textbf{5.81} & 6.87 & \textbf{7.01} & 6.97 & 7.58 & 7.67 &\textbf{ 8.01} \\
    \midrule
     & \multicolumn{15}{c}{\textbf{bn$\rightarrow$en}}\\ 
     \midrule
    char &  3.39 & 3.42 & \textbf{3.58} & 4.03 & 4.12 & \textbf{\textit{4.37}}&4.98 & 5.04 & \textbf{\textit{5.15}}& 5.76 & 5.85 &\textbf{ 5.91} &7.27 & 7.38 &\textbf{ 7.39}\\
    4K & 3.68 & 3.79& \textbf{\textit{4.23}} & 3.15 & 3.66 & \textbf{3.98} & 4.36 & 4.68 & \textbf{4.92} & 6.33 & 6.56 & \textbf{\textit{6.67}} & 7.02 & 7.13 & \textbf{\textit{7.45}} \\
    8K & 3.75 & 3.86 &\textbf{ 3.95} & 3.10 & 3.54 & \textbf{3.76 }& 4.48 & 4.55 & \textbf{4.72} & 5.95 & 6.02 & \textbf{6.12} & 6.64 & 6.83 & \textbf{7.02} \\
    16K & 3.77 & 3.83 & \textbf{3.99} & 3.02 & 3.51 & \textbf{3.65} & 4.31 & 4.48 & \textbf{4.65} & 5.86 & 5.98 & \textbf{6.03} & 6.54 & 6.77 &\textbf{ 6.99}\\
    32K & 3.76 & 3.91 & \textbf{3.93} & 3.14  & 3.42 & \textbf{3.68} & 4.43 & 4.56 & \textbf{4.82} & 5.81 & 5.90 & \textbf{5.93} & 6.83 & \textbf{7.02} & 6.95 \\
    \bottomrule
  \end{tabular}
  \caption{BLEU scores for few-shot performance on transliterated English-Bengali and English-Marathi pairs using character tokenisation and shared BPE with vocabulary size $v$ in $\{4000, 8000, 16000, 32000\}$. Bold shows best score for each vocabulary size and bold italic represents best score overall. }
  \label{tab:fewshot-results}
\end{table*}


\end{document}